\documentclass{agujournal2019}
\usepackage{url} 
\usepackage{lineno}
\usepackage{setspace}
\usepackage{soul}

\usepackage{amssymb}
\usepackage{lmodern}
\usepackage{amsmath, bm}
\usepackage{siunitx}

\usepackage{ragged2e}

\draftfalse

\journalname{Earth and Space Science}

\begin{document}
\justifying

\title{A Deep Generative Model for the Simulation of Discrete Karst Networks}

\authors{Dany Lauzon\affil{1,2}, Julien Straubhaar\affil{2}\thanks{Corresponding Author}, and Philippe Renard \affil{2}}

\affiliation{1}{Department of Civil, Geological and Mining Engineering, Polytechnique Montréal, P.O. Box 6079, Station Centre-Ville, Montréal, H3C 3A7, Québec, Canada}
\affiliation{2}{The Centre for Hydrogeology and Geothermics, University of Neuchâtel, 11 Rue Emile Argand, 2000, Neuchâtel, Switzerland}

\correspondingauthor{Julien Straubhaar}{julien.straubhaar@unine.ch}

\begin{keypoints}
\item Deep generative neural networks are developed to model discrete karstic networks
\item The topology of the karst system is captured using a Graph Recurrent Neural Network
\item Coordinates and karst properties are generated using a Graph Denoising Diffusion Probabilistic Model
\end{keypoints}

\begin{abstract}
The simulation of discrete karst networks presents a significant challenge due to the complexity of the physicochemical processes occurring within various geological and hydrogeological contexts over extended periods. This complex interplay leads to a wide variety of karst network patterns, each intricately linked to specific hydrogeological conditions. We explore a novel approach that represents karst networks as graphs and applies graph generative models (deep learning techniques) to capture the intricate nature of karst environments. In this representation, nodes retain spatial information and properties, while edges signify connections between nodes. Our generative process consists of two main steps. First, we utilize graph recurrent neural networks (GraphRNN) to learn the topological distribution of karst networks. GraphRNN decomposes the graph simulation into a sequential generation of nodes and edges, informed by previously generated structures. Second, we employ denoising diffusion probabilistic models on graphs (G-DDPM) to learn node features (spatial coordinates and other properties). G-DDPMs enable the generation of nodes features on the graphs produced by the GraphRNN that adhere to the learned statistical properties by sampling from the derived probability distribution, ensuring that the generated graphs are realistic and capture the essential features of the original data. We test our approach using real-world karst networks and compare generated subgraphs with actual subgraphs from the database, by using geometry and topology metrics. Our methodology allows stochastic simulation of discrete karst networks across various types of formations, a useful tool for studying the behavior of physical processes such as flow and transport.
\end{abstract}

\section{Introduction}
\label{Sec.Intro}

Karst systems form intricate networks through the dissolution of carbonate rocks. Climate change, marked by extreme weather events such as intense rainfall, has a profound impact on these environments. Due to their complex structure, karst systems react quickly to these events, resulting in the rapid migration of contaminants. Understanding their behavior under such extreme conditions is crucial for managing water quality and mitigating environmental risks. Consequently, modeling karst networks cannot rely on deterministic approaches, as the network strongly influences groundwater flow and transport predictions. A simple estimation of conduit maps fails to capture the full range of possible behaviors in the system, making probabilistic or stochastic modeling essential for more accurate predictions \cite{Chiles2012}.

The methodology for stochastic discrete karst network (DKN) modeling \cite{Jaquet2004, Borghi2012, Fandel2022, Gouy2024, Miville2025} is of the utmost importance for groundwater flow models, because the uncertainty in network structures has a significantly greater impact on prediction than the uncertainty in hydraulic parameters \cite{Refsgaard2006, Fandel2021}. Furthermore, karst formation is driven by physicochemical processes occurring in diverse geological and hydrogeological environments over extended time periods \cite{Palmer1991, Filipponi2009}, making the modeling task difficult. Chemistry-based models are hindered by long computational run times. Fast-marching methods requires to model in the geological domain the velocity propagation field reflecting the hydrogeological properties. Such methods require the specification of numerous parameters, leading to complex setups and complicating the modeling of karst aquifers.

To overcome the complexity of karst formation process and to get rid of a background grid storing geological and hydrogeological properties, we propose to develop a fully machine learning based approach to generate stochastic simulations of karst networks. The idea is to learn the statistics of existing karst conduit networks from datasets that capture different geometries and topologies \cite{Collon2017}, without modeling any physical process. In this context, a karst network is naturally modeled by a graph, whose nodes represent survey points and the edges connect the points.

A key point when representing a karst conduit network as a graph, is that the spatial location of the nodes is crucial, contrary to a graph modeling for example a social community where only the topology of the graph matters, a graph node representing a person and a graph edge a relation between two persons. In this work, the nodes store property information, and the edges represent conduits connecting the nodes. The features, attached to a node, include at least its spatial coordinates, plus any other measure such as the cross-section width and height, which can be derived from the distances to the surrounding walls. The graph-based approach facilitates the exploration of the karst networks distribution, with respect to the connection between the conduits (i.e. topology), the geometry of the network (length and orientation of the edges), and other node properties. However, modeling the complex distributions of karst systems as graphs and efficiently sampling from them presents significant challenges. These arise from limited observational data, the high-dimensional nature of karst networks, and the intricate nonlocal dependencies between edges within a given network.

On one hand, \citeA{You2018} showed that graph recurrent neural networks can approximate a wide range of distributions of graph topologies with minimal assumptions about their structure. Graph Recurrent Neural Network (GraphRNN) \cite{You2018}, or Graph Recurrent Attention Networks \cite{Liao2019} are approaches capable to learn the topology of graphs with different number of nodes, and can be used to stochastically generate new graphs. On the other hand, \citeA{Simonovsky2018} developed a methodology based on graph variational auto-encoders, accounting for features, to generate small graphs representing molecules. Furthermore, generative diffusion models \cite{Ho2020}, and in particular graph diffusion models \cite{Zhu2022, Zhang2023} can be used to learn the geometry and associated features of graphs and to stochastically simulate new graphs \cite{Hoogeboom2022, Vignac2022, Zhang2023, Igashov2024}. These approaches are primarily used to analyze social network interaction, to predict traffic patterns, or to discover new molecules, proteins, and drugs, along with their properties. However, these methods have not yet been utilized to predict complex geological mechanisms, particularly in the context of karstic formations.

Therefore, we explored the use of deep generative models to generate stochastic DKNs, integrating topological and feature-based predictions separately. Our implementation consists of a two-step process: (1) predicting the karst network's topology using GraphRNN and (2) generating geometry and features through Graph-Denoising Diffusion Probabilistic Models (G-DDPM). Each network is independently trained on a dataset of karst networks. The generation of new graphs is done by sampling a graph topology through the GraphRNN, then using it as input for the G-DDPM model that is employed to sample the features attached to the graph nodes. This allows for the stochastic simulation of karst networks that accurately capture structural characteristics while also providing estimates of geological features.

The structure of the paper is as follows. We begin by introducing the two deep learning algorithms, GraphRNN and G-DDPM, which are used to stochastically simulate karst networks through deep generative graph models. Next, we outline the architecture of our networks and the key parameters selected. We then introduce a set of statistical metrics to evaluate and compare the generated graphs with the original networks. Following this, to demonstrate the capabilites of our methodology, we apply it to two different real-world karst systems, the first one being essentially bi-dimensional (i.e. karst formation in a plane) and the second one depicting three-dimensional patterns with attached information about the size of the aperture of the conduits. Finally, we conclude with a discussion of the limitations of our approach and explore potential directions for future research.

\section{Methodology}
\label{Sec.Metho}

Before tackling the problem of stochastically simulating DKNs using deep generative models, it is essential to understand the fundamental concepts of the two main algorithms (GraphRNN and G-DDPM) used in the two steps of the proposed approach.

\subsection{Graph Recurrent Neural Networks}
\label{Sec.GraphRNN}

The objective of a graph generative model is to learn the distribution \( p_{\text{m}}(G) \) over graphs using a collection of observed graphs \( \mathbb{G} = \{ G_1, \ldots, G_s \} \) sampled from a data distribution \( p(G) \). Each graph \( G_i \) in the dataset may vary in the number of nodes and edges it contains. In this paper, a graph \( G = (V, E) \) is considered undirected, defined by its set of vertices \( V = \{ v_1, \ldots, v_n \} \), referred to as nodes and its set of edges \( E \subset \{ \{v_i, v_j\} \mid v_i, v_j \in V \} \), an edge \(\{u, v\}\) being an unordered pair of vertices meaning that the nodes \(u\) and \(v\) are connected.

GraphRNN, introduced by \citeA{You2018}, is a deep autoregressive model that breaks down the graph generation process into sequential steps for the creation of nodes and edges, based on the current structure of the graph. Using a sequence-to-sequence framework based on recurrent neural networks (RNNs), it models the generation of nodes and edges as a sequence task. The output of each step depends on its preceding values and incorporates a stochastic component. Consequently, each step corresponds to the addition of a node or edge to the evolving graph.

\begin{figure}[!ht]
    \centering
    \includegraphics[width=0.95\linewidth]{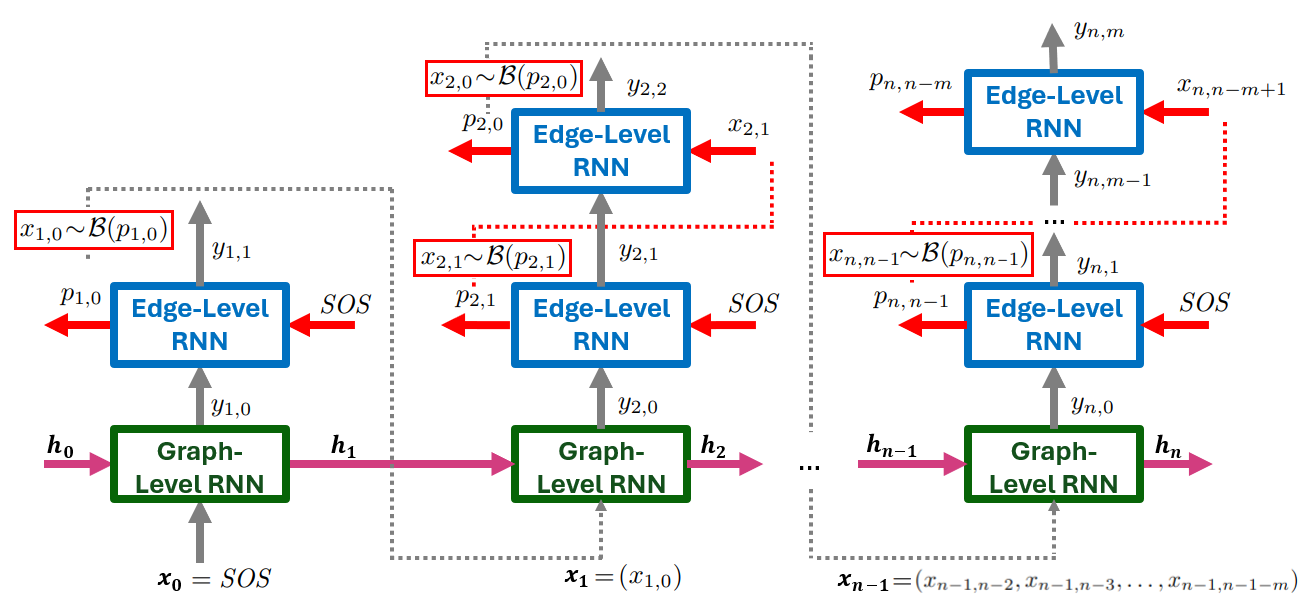}
    \caption{Illustration of GraphRNN (\( \bm{x} \): input vector; \( \bm{h} \): hidden state vector; \( \bm{y} \): output vector; \( \bm{p} \): vector of predicted probabilities; \textit{SOS}: start-of-sequence, a vector of ones; \(\mathcal{B}(p)\): Bernoulli distribution of parameter \(p\)).}
    \label{fig.GraphRNN}
\end{figure}

GraphRNN is based on two RNN units: one for predicting node generation (Graph-Level RNN) and the other for generating connections between nodes (Edge-Level RNN). Fig.~\ref{fig.GraphRNN} illustrates the working process of the Graph-Level and Edge-Level RNNs. In each iteration, the Graph-Level RNN attempts to answer the question ``Do we need to generate another node~?'', while the Edge-Level RNN responds sequentially to the question ``To which nodes generated so far do we need to connect the newest node~?''

The process depicted in Fig.~\ref{fig.GraphRNN} works as follows. Initially, a graph with only one node, \(v_0\), is considered. In iteration 1 (left part in the figure), the Graph-Level RNN generates latent features \( y_{1,0} \), and passes this information to the Edge-Level RNN. The Edge-Level RNN, based on \( y_{1,0} \), predicts a probability, \( p_{1,0} \) of connecting node \(v_1\) to node \(v_0\). Then, a Bernoulli distribution with probability \( p_{1,0} \), \(\mathcal{B}(p_{1,0})\), is sampled to decide whether to create the edge. If the sampled value is \(x_{1,0}=1\), a new node \(v_1\) is created and connected to the node \(v_0\). In iteration 2 (middle part in the figure), the Graph-Level RNN generates latent features \( y_{2,0} \), conditioned on what has been generated so far: the sampled vector \(x_1\) and the hidden state \(h_1\). Probabilities \( p_{2,1} \) and \( p_{2,0} \) are then sequentially predicted through the Edge-Level RNN, \( p_{2,i} \) being the probability to connect node \(v_2\) to node \(v_i\). Note that \(x_{2,1}\sim\mathcal{B}(p_{2,1}) \) is sampled first, then passed to the Edge-Level RNN to predict \( p_{2,0} \), and next, \(x_{2, 0}\sim\mathcal{B}(p_{2,0}) \) is sampled. The sampled values \(x_{2,i}\) indicates the presence (1) or absence (0) of an edge linking nodes \(v_2\) and \(v_i\). Thus, the process is purely sequential. The sampled values are concatenated in a vector \(x_2=(x_{2,1}, x_{2,0})\) that feeds the Graph-Level RNN in the next iteration. This procedure is repeated until either the vector \(x\) is null, i.e. the next node is unconnected (no new edge) or a maximum number of nodes \( n_{max} \) is reached.

Note that for a new node \(v_n\), the possible connections are restricted to the \(m\) previous nodes (see the right part in Fig.~\ref{fig.GraphRNN}). This allows to limit the size of the vectors \(x\), and enhance the performance of the algorithm. Practically, all the vectors \(\bm{x}_1, \bm{x}_2,\ldots\) are of length $m$ filled with zeros in the first iterations (See Sec.~\ref{subsubsec.AdjEncode} for the definition of parameter $m$). A start-of-sequence (SOS) is a vector of ones, to activate the network.

Of course, this approach depends on the node ordering, this is discussed in the next section.

\subsubsection{Node Ordering and Adjacency Matrix Encoding}
\label{subsubsec.AdjEncode}
One significant challenge with a sequential approach is that every node must be visited to ensure all possible connections, potentially leading to time-consuming edge generation steps. To tackle this, GraphRNN employs a Breadth-First Search (BFS) node ordering strategy, which restricts the sequential approach to a specific neighborhood of length \( m \), which drastically improve scalability \cite{You2018}. For each graph, the nodes are ordered so that any newly added node connects primarily to the most recently added nodes. This approach limits the number of preceding nodes considered during the generation process by restricting edge generation to the \( m \) previous nodes, where \( m \ll n \) in most cases.

Basically, for a given graph with \(n\) nodes, numbering the nodes according to a BFS strategy consists in first randomizing the nodes numbers (\(0, 1, \ldots, n-1\)), then building the BFS sequence of visited nodes: 1) start with \(B\!F\!S = [0]\) (i.e. start with node 0), and 2) repeat until all the nodes have been visited: find the first node number in \(B\!F\!S\) that has at least one neighbor (i.e. node connected with an edge) not yet in \(B\!F\!S\), and append all its neighbors to \(B\!F\!S\).

\begin{figure}[!ht]
    \centering
    \includegraphics[width=0.80\linewidth]{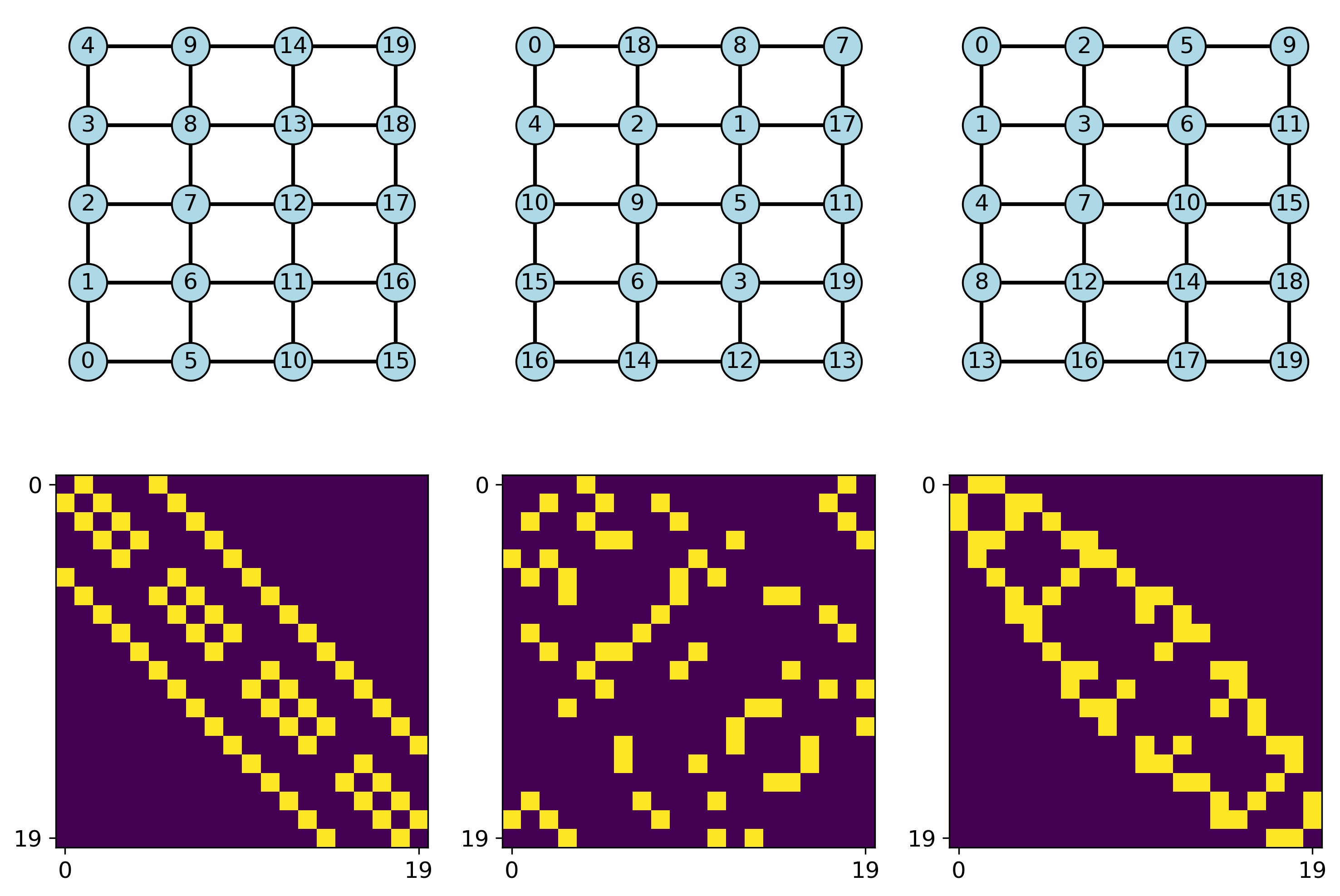}
    \caption{Different node orderings \( \pi \) on a 5 \( \times \) 4 grid graph. Left: Numbering from bottom to top and left to right. Middle: Random ordering. Right: one BFS ordering.}
    \label{fig.BFSOrdering}
\end{figure}

Once an ordering \(\pi\) of the nodes is specified for a graph with \(n\) nodes, it can be represented by an adjacency matrix \(A = (a_{i,j})_{0\leq i, j <n}\), a binary squared matrix of order \(n\) with \(a_{i,j}\) equals one if the nodes \(i\) and \(j\) are connected by an edge, and equals zero otherwise. Note that for undirected graphs, the adjacency matrix is symmetric. Three different orderings are illustrated in Fig.~\ref{fig.BFSOrdering} for a ``grid'' graph. Using a random order (middle), Edge-Level RNN should visit each previous node to evaluate the probability of connecting the newest node to others (e.g., \( p_{i,j}\) has to be predicted for all \(i < j\), \(j=1,\ldots, n-1\)). In contrast, the BFS sequence (right) arranges the nodes ordering so that the connections are centered around the diagonal. This means that for a given distance \( m \), it is certain that the probability of a connection outside this region is zero (e.g., \( p_{i,j} = 0, \) if \( i - j > m \)). It is also possible to have a node ordering \( \pi \) that does not follow a BFS sequence (left) and arranges the node connections around the diagonal. However, such behavior is not useful from a mathematical point of view for our application.

\begin{figure}[!ht]
    \centering
    \includegraphics[width=0.80\linewidth]{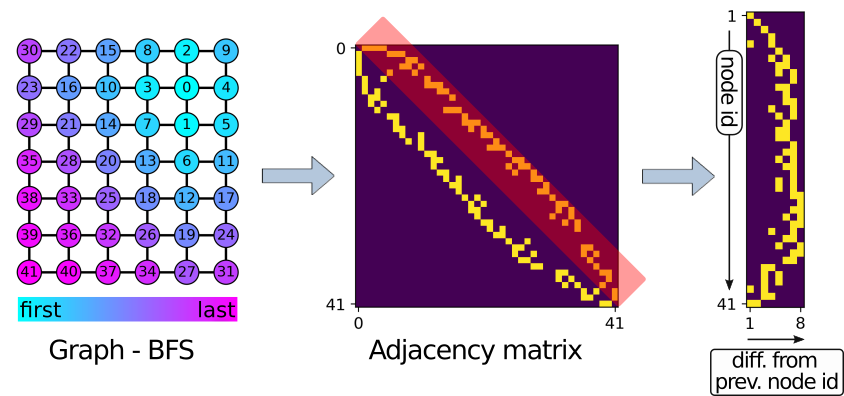}
    \caption{Procedure for embedding a graph structure into a matrix using a BFS sequence. Parameter \( m\) equal 8.}
    \label{fig.BFSEmbedding}
\end{figure}

Once a BFS ordering is performed, and using the symmetry of the adjacency matrix, this latter can be embedded to reduce its dimensionality, by storing only the \(m\) upper diagonals (above the main one) (Fig.~\ref{fig.BFSEmbedding}). The resulting matrix (Fig.~\ref{fig.BFSEmbedding} right), \(C = (c_{i,j})_{1\leq i < n, 1\leq j \leq m}\), of dimension \(n-1 \times m\), is defined as \(c_{i, j} = a_{i-j, i}\) (if \(i-j\geq 0\) and \(0\) otherwise. It is important to note that this embedding is reversible and corresponds to a unique graph or adjacency matrix, which means that the transformation preserves a one-to-one correspondence between the graph and the embedded matrix. Note that the rows of the encoded matrix (at right in Fig.~\ref{fig.BFSEmbedding}) correspond to the vectors \(x\) given as inputs to the Graph-Level RNN in Fig.~\ref{fig.GraphRNN}.

To estimate the parameter \( m \), a large number (e.g. \(\num{100000}\)) of BFS node orderings are performed on graphs sampled from the observed set \( \mathbb{G} = \{ G_1, \dots, G_s \} \). Note that, as each graph \( G_i \) have several BFS node orderings (which depends on the initial random permutation of the node numbers), the cardinality of \( \mathbb{G} \) may be less than the number of BFS orderings. Then, the embedding is applied to evaluate \( m \) for each sampled graph with its corresponding BFS node ordering and the parameter \( m \) is set as approximately the 99.9\textsuperscript{th} percentile to capture the majority of the dimensionalities. In general, graphs with regular structures tend to have smaller \( m \), while random or community graphs tend to have larger \( m \) \cite{You2018}. As karst networks are particularly regular, small values of \( m \) are expected.

\subsubsection{Recurrent Neural Network}

A recurrent neural network (RNN) is a deep learning model designed to analyze and transform sequential data into a corresponding sequential output. As mentioned previously, GraphRNN is built around two RNNs (Graph-Level RNN and Edge-Level-RNN). In the implementation of the proposed method, both employ Gated Recurrent Units (GRU).

Introduced by \citeA{Cho2014}, GRU was developed to address the problem of vanishing gradients commonly encountered in standard RNN. To mitigate this issue, the GRU introduces two gating mechanisms: the update gate and the reset gate. These gates are vectors that control which information should be retained or discarded as the data progresses through the network. Their specificity lies in their ability to be trained to retain important information from earlier in the sequence while discarding irrelevant data, thus improving the model's accuracy in capturing long-term dependencies \cite{Cho2014}.

A GRU cell consists of the following operations at timestep \(t\):

\begin{equation}
\begin{aligned}
    z_t &= \sigma (W_z \cdot x_{t-1} + U_z \cdot h_{t-1} + b_z)   \\
    r_t &= \sigma (W_r \cdot x_{t-1} + U_r \cdot h_{t-1} + b_r)   \\
    \hat{h}_t &= \phi (W_h \cdot x_{t-1} + U_h \cdot (r_t \odot h_{t-1}) + b_h) \\
    h_t &= z_t \odot h_{t-1} + (1 - z_t) \odot \hat{h}_t
\end{aligned}
\end{equation}

\noindent
where \( x_{t-1} \in \mathbb{R}^d \) is the input vector, \( h_{t-1} \in \mathbb{R}^e \) the input hidden vector, \(h_t \in\mathbb{R}^E\) the output (hidden) vector, \( \hat{h}_t \in \mathbb{R}^d \) is a candidate vector, \( z_t \in (0,1)^e \) and \( r_t \in (0,1)^e \) are respectively the update and reset gate vectors, \( W_z, W_r, W_h \in \mathbb{R}^{e \times d}, U_z, U_r, U_h \in \mathbb{R}^{e \times e}, b_z, b_r, b_h \in \mathbb{R}^e \) are matrices and vectors of parameters that are learned during training, \( \sigma \) is the logistic activation function (namely the sigmoid function), \(\phi \) is the hyperbolic tangent activation function (namely the tanh function), and \(\odot\) denotes the Hadamard product (i.e. element-wise product). It is important to note that the weight matrices \(W_*\) and the vectors \(U_*\) and \(b_*\) do not depend on the timestep.

Multiple GRU layers can be stacked on top of each other, each layer having its own set of parameter matrices and vectors. The output of one layer serves as the input of the next layer. This architecture enhances the network’s ability to learn complex features and long-range dependencies in sequential data. The goal of such a model is to enable each recurrent layer to capture different scales of dependencies.

The architecture used in our implementation is described in detail in Sec.~\ref{Sec.GraphRNN_architecture}.

\subsubsection{Training GraphRNN}

Training both the Graph-Level and Edge-Level RNNs is done by ``teacher forcing''. This involves using the real previous output (e.g., ground truth) from the training data as input for the next time step in a sequence model, rather than depending on the model's predictions. That is, the real binary values \(e_{n, n-i}\), for presence or absence of edge between node \(n\) and node \(n-i\), taken from the graphs of the dataset are considered instead of the sampled values \(x_{n, n-i}\sim\mathcal{B}(p_{n,n-i})\) to feed each step of the GraphRNN (see Fig.~\ref{fig.GraphRNN}). This approach improves learning by providing the model with precise context at every stage, resulting in better convergence and reducing the accumulation of errors. At inference time, the model generates a graph using its predictions at each time step. The parameters of the network are trained by applying a gradient descent to the loss defined as the binary cross-entropy \cite{Goodman1991} of the real values \(e_{n, n-i}\) and the predicted probabilities \(p_{n, n-i}\).

\subsection{Denoising Diffusion Probabilistic Models}
\label{Sec.DDPM}

Denoising Diffusion Probabilistic Models (DDPM) are a class of generative models that operate by progressively adding noise to an input signal, such as an image, text, audio, or graph, and then learning to reverse this noise, ultimately generating new samples through the denoising process \cite{Ho2020}. DDPM constructs two parameterized Markov chains to diffuse the data with predefined noise and to reconstruct the desired samples from the noise. The model consists of a forward chain for adding noise and a denoising chain for reconstructing samples from noisy structures. Notably, the application of this approach to graphs referred to as G-DDPM, follows a similar methodology, treating graphs as if they were images.

\subsubsection{Forward process}

 In the forward chain, G-DDPM gradually adds noise to the raw data distribution of the graph \( \bm{x}_0 \sim q(\bm{x}_0) \), converging to a multivariate Gaussian distribution \( \bm{x}_T \), where all the variables are uncorrelated (and therefore independent). This implies that the off-diagonal elements of the covariance matrix are zero, meaning there is no linear relationship between the variables, or element on the perturb graph after \( T \) iterations. This process is performed under a pre-designed mechanism. Mathematically, the definition of the forward process \( q \) is as follows :

\begin{equation} \label{Eq.DDPM_ForwardPass_1}
\begin{aligned}
q(\bm{x}_t | \bm{x}_{t-1}) &= \mathcal{N} (\bm{x}_t; \sqrt{1-\beta_t}\bm{x}_{t-1}, \beta_t \bm{I} ) \\
q(\bm{x}_{1:T} | \bm{x}_{0}) &= \prod_{t=1}^T q(\bm{x}_t | \bm{x}_{t-1})
\end{aligned}
\end{equation}

\noindent
where \( \beta_t \in (0,1) \) represents the variance of the Gaussian noise added at timestep \( t \).

One can define a set of equations that enable one to sample the noised latent variable \( \bm{x}_t \) at an arbitrary timestep \(t\) conditioned on \( \bm{x}_0 \) by posing  \( \alpha_t = 1 - \beta_t \), and \( \Bar{\alpha}_t = \prod_{i=1}^T \alpha_i\). Therefore Eq.~\eqref{Eq.DDPM_ForwardPass_1} can be rewritten as \cite{Ho2020} : 

\begin{equation} \label{Eq.DDPM_ForwardPass_2}
\begin{aligned}
q(\bm{x}_t | \bm{x}_0) &= \mathcal{N} (\bm{x}_t; \sqrt{\Bar{\alpha}_t}\bm{x}_{0}, (1-\Bar{\alpha}_t) \bm{I} ) \\
 \bm{x}_{t} &= \sqrt{\Bar{\alpha}_t}\bm{x}_{0} + \sqrt{(1-\Bar{\alpha}_t)} \bm{\epsilon}_t
\end{aligned}
\end{equation}

\noindent
where \( \bm{\epsilon_t}\sim\mathcal{N}(0, \bm{I}) \) denotes the Gaussian noise.

This process is entirely automatic and does not involve any learning phase. In our implementation, the vector \(\bm{x}_0\) represents all the features (coordinates, geometry attributes, etc.) attached to all nodes of a graph. Hence, the forward process simply consists in adding noise to the graph node features of any graph in the considered database \(\mathbb{G} = \{ G_1, \ldots, G_s \} \) to create a dataset used to learn the denoising process. The noise is sequentially introduced through timesteps \( t \) varying from \( 1 \) to \( T \), according to a predefined noise schedule \(\beta_1, \ldots, \beta_T\).

\subsubsection{Denoising process}

The second Markov chain is called the denoising process (or reverse process). It aims to learn a transition distribution \( p_{\theta} \) parameterized by \( \theta \) that takes noisy data as input and predicts the clean data. The equation can be written as:

\begin{equation} \label{Eq.DDPM_ForwardPass_3}
\begin{aligned}
p_{\theta}(\bm{x}_{0:T}) &= p(\bm{x}_T) \prod_{t=1}^T p_{\theta}( \bm{x}_{t-1} | \bm{x}_t ) \\
p_{\theta}( \bm{x}_{t-1} | \bm{x}_t ) &= \mathcal{N} \left( \bm{x}_{t-1}; \mu_{\theta}(\bm{x}_t,t), \Sigma_{\theta}(\bm{x}_t,t) \right)
\end{aligned}
\end{equation}

This reverse process needs to be learned to produce new observations of a graph. Therefore, a deep neural network is used to learn the parameterization \( \theta \); note that in practice the variance \(\Sigma_{\theta}(\bm{x}_t,t)\) in the model is often set to \(\sigma_t \bm{I}\), with e.g. \(\sigma_t = \beta_t\), that is the variance is not learned. The architecture can be complex to determine. In our implementation, we decided to explore the option of fixing the graph topology while applying noise solely to the graph node features. This results in a purely continuous deformation, without categorical data, i.e., the node features consist of continuous variable reflecting coordinates and speleological survey data.

Note that during training, the graph topologies of the database are used, whereas during generation, new plausible features must be recovered (through the denoising process) on the output graphs of the GraphRNN.

\subsubsection{Deep neural network} \label{Sec.GDDPM_general}

A classical way to learn the denoising process is to use a ``re-parametrization trick'' \cite{Ho2020} that leads to learn the noise added to the initial features \( \bm{x}_{0} \) to get \(\bm{x}_{t}\) according to Eq.~\eqref{Eq.DDPM_ForwardPass_2}. That is, for any timestep \(t\), a deep neural network is used to learn \(\epsilon_t\) from \(\bm{x}_{t} = \sqrt{\Bar{\alpha}_t}\bm{x}_{0} + \sqrt{(1-\Bar{\alpha}_t)} \bm{\epsilon}_t\). In other words, the deep neural network learns the mapping \( \bm{x}_t \mapsto \bm{\epsilon}_t\), for any \(t\) in \(\{1,\ldots, T\}\).

Similar to diffusion models for images, which apply noise independently to each pixel, our approach involves diffusing noise separately onto each feature of a graph. Considering images, the deep neural network learns image-to-image mappings above, where \(\bm{x}_t\) and \( \bm{\epsilon}_t\) are features (channels colors) on each pixel. In our approach, graphs are considered and the deep neural network learns graph-to-graph mappings where \(\bm{x}_t\) and \( \bm{\epsilon}_t\) are features attached to the graph nodes. As the edges are predetermined, the topology of the graph is fixed, and only the continuous features remain to be processed. Note that although only node features are considered in our approach, extending it to edge features is straightforward.

Considering a network per timestep would be inefficient because the total number of parameters to learn would be enormous. To tackle this, a single deep neural network is designed for an image-to-image or graph-to-graph mapping and a sinusoidal position embedding \cite{Vaswani2023} is added to account for the timestep.

U-Net \cite{Ronneberger2015} is a traditional architecture to learn image-to-image mappings, involving Convolutional Neural Networks (CNNs). The deep neural network used in our implementation, for learning graph-to-graph mappings, is inspired by the U-Net design and the CNNs are replaced by Graph Convolutional Networks (GNNs). See Sec.~\ref{Sec.GDDPM_network} for details.

Finally, the Mean Square Error (MSE) loss function is used to train the network.

\section{Deep Learning Architectures, Model Hyperparameters}
In this section, the architectures and hyperparameters of the networks used in our implementation of GraphRNN and G-DDPM are described.

\subsection{GraphRNN architecture}  \label{Sec.GraphRNN_architecture}

\begin{figure}[!ht]
    \centering
    \includegraphics[width=0.65\linewidth]{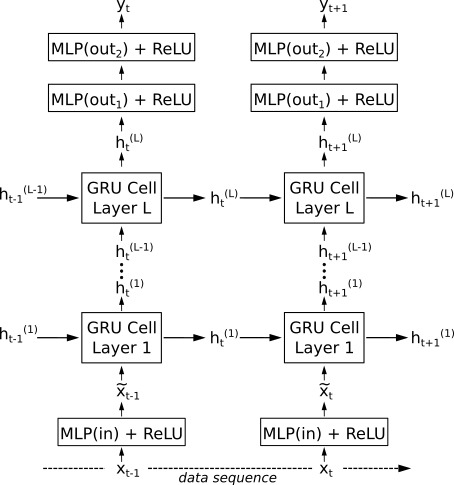}
    \caption{Design of the RNN model (GRU: Gated Recurrent Unit, MLP: Multi-Layer Perceptron, ReLU: Rectified Linear Unit).}
    \label{fig.RNN_model}
\end{figure}

The design of Graph-Level RNN and Edge-Level RNN, defined in Sec.~\ref{Sec.GraphRNN}, are both based on GRU Cells with embedding Multi-Layer Perceptrons (MLPs) at input and output as described in Fig.~\ref{fig.RNN_model}.

The Graph-Level RNN employs \(L_G=4\) layers of GRU cells, with a \(d_{G, hidden}=48\)-dimensional hidden state in each layer (vectors \(h\) in Fig.~\ref{fig.RNN_model}). The input dimension (vector \(x\)) is \( d_{G, in} = m \), where \(m\) is the parameter defined in Sec.~\ref{subsubsec.AdjEncode}. The input embedding MLP (MLP(in)) maps a \(d_{G, in}\)-dimensional space to a \(d_{G, G\!R\!U(in)}=64\)-dimensional space (vector \(\tilde x\)). The first output MLP (MLP(out\(_1\)) goes from dimensions \(d_{G, hidden}\) to \(d_{G, out, 1}\) and the second output MLP (MLP(out\(_2\)) from dimensions \(d_{G, out, 1}=16\) to \(d_{G, out, 2}=32\) (vector \(y\)). Note that each MLP is followed by a Rectified Linear Unit (ReLU) activation.

The Edge-Level RNN uses a similar structure with the following hyperparameters. The number of layers is also set to \(L_E=4\). The input dimension and the output dimension are \(d_{E, in} = d_{E, out, 2} = 1\), a single number representing the (probability of) connection between two graph nodes. The hidden state is of dimension \(d_{E, hidden} = d_{G, out, 2}\), which is imposed by the architecture of the GraphRNN. The input dimension of the first layer of GRU cell is set to \(d_{E, G\!R\!U(in)}=24\), and the output dimension of the first output MLP (MLP(out\(_1\)) to \(d_{E, out, 1}=36\).

Using the same RNN model as depicted in Fig.~\ref{fig.RNN_model} for Graph-Level RNN and Edge-Level RNN facilitates the implementation. This is not a limitation, since the hyperparameters \(L_G, d_{G, hidden}, d_{G, G\!R\!U(in)}, d_{G, out, 1}, d_{G, out, 2}\) and \(L_E, d_{E, G\!R\!U(in)}, d_{E, out, 1}\) can be set independently, which offers high flexibility. Note that the selected hyperparameters described above have been set after some trial, but other setups could be chosen.

\subsection{G-DDPM network} \label{Sec.GDDPM_network}

\begin{figure}[!ht]
    \centering
    \includegraphics[width=0.95\linewidth]{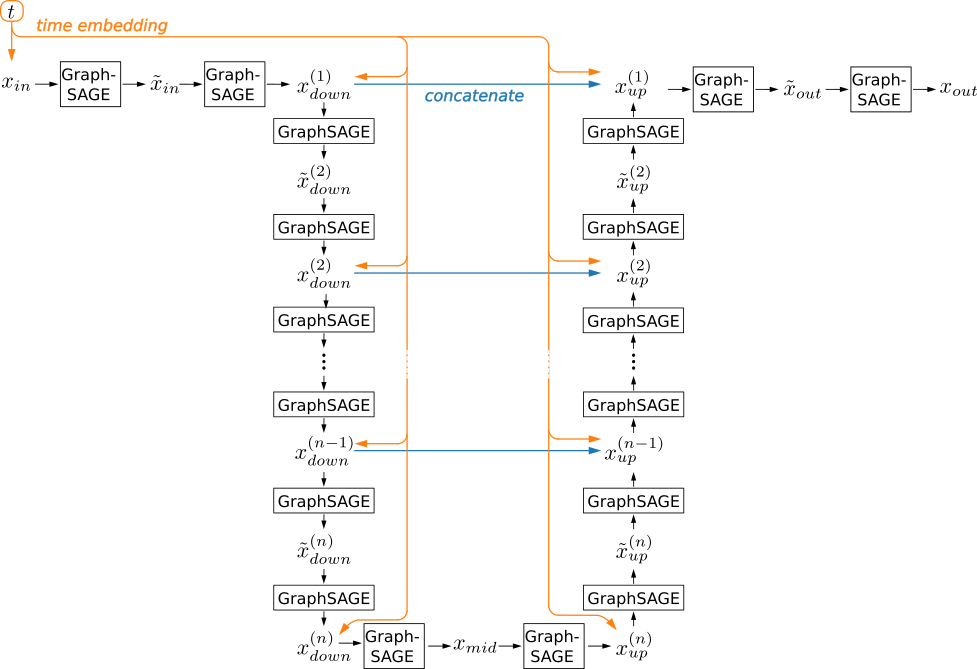}
    \caption{Design of the GDDPM: U-net architecture, based on GraphSAGE operations and time embedding.}
    \label{Fig.GDDPM_Unet}
\end{figure}

The G-DDPM is a graph neural network designed to learn the graph-to-graph mappings \( \bm{x}_t \mapsto \bm{\epsilon}_t\), for \(t= 1,\ldots, T\), where \(\bm{x}_t\) the noisy graph node features at timestep \(t\), and \(\bm{\epsilon}_t\) the Gaussian noise added from the original features, see Sec.~\ref{Sec.GDDPM_general}. A U-Net-inspired architecture \cite{Ronneberger2015} is used, as shown in Fig.~\ref{Fig.GDDPM_Unet}, and explained below.

\begin{figure}[!ht]
    \centering
    \includegraphics[width=0.80\linewidth]{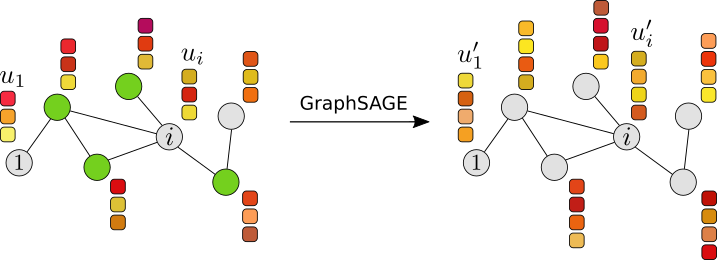}
    \caption{Illustration of GraphSAGE operation: \(u_i\) is the input vector of features attached to node \(i\), the green nodes are the neighbors \(\mathcal{N}(i)\) of the node \(i\) involved in a GraphSAGE operation: \(u_i' = W_1 \cdot u_i + W_2 \cdot \operatorname{mean}_{j\in\mathcal{N}(i)} u_j\) ; input and output vectors of features are of dimensions \(d_{in}=3\) and \(d_{out}=4\) on this example.}
    \label{fig.GraphSAGE}
\end{figure}

The network takes as input the set of input features \(x_{in} =\bm{x}_t\) defined on a connected graph, and the corresponding timestep \(t\), and predicts the noise \( \hat{\bm{\epsilon}}_t \) through a sequence of operations. The U-Net includes downward operations, a middle operation (also known as bottelneck), and upward operations that involve concatenation with corresponding features from the downward path. Unlike the U-Net architecture for images, which relies on CNNs and changes the image scale during processing, the proposed method, applied to graphs, maintains the same graph structure throughout the entire process using GraphSAGE layers. A GraphSAGE operation (Fig.~\ref{fig.GraphSAGE}) allows to aggregate information from node attributes and their neighbors: if \(u_i\) denotes the vector of input features of dimension \(d_{in}\) attached to the node \( i \), the output features is a vector \(u_i'\) of dimension \(d_{out}\) defined as

\begin{equation}
u_i' = W_1 \cdot u_i + W_2 \cdot \operatorname{mean}_{j\in\mathcal{N}(i)} u_j
\end{equation}

\noindent
where \(\mathcal{N}(i)\) denotes the neighbors of the node \(i\), i.e. (the nodes \( j\)  linked to node \(i\) by an edge. The parameter matrices \(W_1, W_2 \in \mathbb{R}^{d_{out}\times d_{in}}\) are learned by the network.

Applying \(n\) successive GraphSAGE operations allows to combine information from nodes in a \(n\)-neighborhood, i.e. nodes reached by a path of a maximum of \(n\) edges. In our implementation, the number \(n\) in Fig.~\ref{Fig.GDDPM_Unet} is set to \(n=25\), allowing deep aggregation of information along the edges. The feature vector in input of the U-Net is \(d\), the number of features attached to the graph nodes, which depends on the case study (at least 2 or 3 spatial coordinates + possible geometry features). In the other steps of the U-Net, the feature vector dimension is an hyperparameter set to \(8\cdot d\), but it could be varying along the net. Note that a final linear operation (not shown on the Fig.~\ref{Fig.GDDPM_Unet}) maps the output feature vector to a feature vector of dimension \(d\), to match the input dimension, and to refine the predicted noise.

\section{Statistical metrics of network topology and geometries}
\label{Sec.Metrics}

To evaluate the proposed deep neural network methodology, statistics of topological and geometrical measures \cite{Collon2017}, computed on the graphs of the data set (used for training) and on generated graphs, are compared. Ten metrics are selected and computed using the open-source python package Karstnet (https://github.com/karstnet/karstnet), a Python project that provides tools for the statistical analysis of karst networks. These metrics are presented in Table~\ref{Tab.KarstNetStats} and briefly discussed below, where parentheses refer to the row numbering in the table. The metrics marked with (*) in the table are topological measures that are computed on the simplified graph, obtained by removing nodes of the graph while preserving its topology.

These metrics involves two key notions. The degree of a node is the number of edge(s) attached to that node, and a branch refers to a path connecting two nodes of degree not equal to \(2\), where all intermediate nodes have a degree of \(2\). Given a graph, the simplified graph is obtained by removing all the nodes of degree \(2\) in each branch, but keeping one intermediate node if more than one branch has the same pair of distinct extremities and keeping two intermediate nodes in branches for which the two extremities (endpoints) are the same nodes.

The first five metrics are geometrical measures, which account for the node position (and edge length) of the graphs. The mean (1), the coefficient of variation (2), and the entropy (3) of the branch length are important parameters for describing the geometry of karst networks. The coefficient of variation shows the extent of variability of conduit lengths relative to their mean. The entropy provides information about the distribution of conduit lengths. It is computed over ten bins as \(H=-\sum_{i=1}^{10} p_i \log_{10} p_i\), where \(p_i\) is the proportion of branches having a length in the \(i\)-th bin; a uniform distribution gives an entropy of one, whereas a network whose all the branches have the same length has an entropy of zero. The tortuosity (4) of a branch is defined as the ratio of its length to the distance between its endpoints. It is a classical metric used to characterize karstic morphologies \cite{PardoIguzquiza2011}, and the mean tortuosity of the network is considered. The entropy of the edges' orientation (5) can help detect the existence of geological features: tectonic (joints, fractures, and faults) or stratigraphic (bedding planes); the azimuth angles in a horizontal plane are used, with bins of 10\textsuperscript{o}, to compute the entropy.

The last five metrics are topological measures computed on the simplified graph. The shortest path length between two nodes in a graph is equal to the minimal number of edges used to connect them, and the average shortest path length (6*) is the mean over shortest paths in the graph. This measure is also the inverse of the harmonic mean of the nodes' closeness centrality \cite{freeman1978}. The average shortest path length provides insight into transport efficiency, which is highly correlated with the size of the network. Central point dominance (7*) helps identify the presence of a centralized network organization. The central point dominance is zero for a complete graph, i.e., a graph in which each node has a direct edge to all the other nodes, and one for a star graph in which the same central node is included in all paths; a star graph with \(n\) nodes has exactly \(n-1\) edges connecting one node to all the other ones. More precisely, the central point dominance for a graph \(G\) with \(n\) nodes is expressed as \(cpd = 1/(n-1) \cdot \sum_{v\in G} (B(v^\star) - B(v))\), where \(B(v)=2/((n-1)(n-2))\cdot\sum_{\{s, t\}\subset G, s\neq t, s\neq v, t\neq v} \xi(s,t\,\vert\,v)/\xi(s,t)\), with \(\xi(s,t)\) denoting the number of shortest paths linking \(s\) and \(t)\), and \(\xi(s,t\,\vert\,v)\) the number of these paths going through \(v\), and where \(v^\star=\operatorname{argmax}_{v\in G} B(v)\). The number \(B(v)\) is the betweenness centrality (normalized, i.e. the factor \(2/((n-1)(n-2))\) ensures that \(B(v)\leqslant 1\) for any node in any graph). The central point dominance is the mean of the (normalized) betweenness centrality exceedance of the most central node (according to this measure) over every other node. Node degree correlation (8*) is easy to calculate and is associated with the predominance of assortative networks in karst systems. It computes the correlation of degrees between first-neighbor nodes (i.e. nodes connected by one edge). A positive value indicates an assortative network, i.e., nodes of high degree tend to connect with nodes of high degree, whereas a negative value is associated with disassortative networks, i.e., nodes of high degree tend to connect to nodes of low degree. The mean of node degrees (9*) is particularly useful because it is also simple to compute, effectively distinguishes interconnected systems from tree-like structures, and allows the classification of acyclic systems based on the total number of branches. Finally, the coefficient of variation of node degrees (10*) can be computed to characterize the dispersion of degrees.

For a more detailed explanation of statistical metrics and their use in characterizing karst network geometry and topology, readers are encouraged to refer to \citeA{Collon2017}.

\begin{table}[!ht]
    \centering
    \caption{Selected statistical metrics for the characterization of karst network geometry and topology (*: metric computed on the simplified graph where nodes of degree two are removed, and the graph topology preserved).}
    \begin{tabular}{l|l}
    \hline
        \textbf{Method}  & \textbf{Description} \\ \hline
        (1) `mean length' & Mean length of the branches \\
        (2) `cv length' & Coefficient of variation (std. dev. / mean) of branch length \\
        (3) `length entropy' & Entropy of branch length \\
        (4) `mean tortuosity' & Mean tortuosity of branches \\
        (5) `orientation entropy' & Entropy of edges orientation \\
        (6*) `aspl' & Average shortest path length \\
        (7*) `cpd' & Central point dominance \\
        (8*) `node degree corr.' & Correlation of node degrees over pairs of connected nodes \\
        (9*) `mean degree' & Mean of node degrees \\
        (10*) `cv degree' & Coefficient of variation of node degrees \\
         \hline
    \end{tabular}
    \label{Tab.KarstNetStats}
\end{table}

\section{Application to Real-World Karst Systems}
\label{Sec.Application}

Two real-world mapped karst network are selected to evaluate our approach. The first network represents the underwater cave system Ox Bel Ha \cite{Coke2019}, located in Mexico's Yucatán Peninsula. Due to its unique formation as an underwater coastal system, it can be viewed as a two-dimensional system. The second network corresponds to the karstic network of Sakany \cite{cassou2007}, located in southern France, in the Ariège department (Midi-Pyrénées region), within the Vicdessos River valley. This is a complex three-dimensional system, developed in a karstic massif, with intricate conduit geometries.

For both case studies, the same procedure is applied to build a dataset, train the deep neural networks and generate new graphs. This is detailed in the next section. Then, geological context and the results for each case study are presented with visualizations of the first graphs of the dataset and the first generated graphs, and statistics comparison using the metrics of Sec.~\ref{Sec.Metrics} between the original networks (dataset) and the generated networks (generated set).

\subsection{Dataset and generation of new graphs}

One main graph representing the mapped karst network is given. Training machine learning algorithms requires a large amount of data. Hence, we propose a data augmentation scheme that samples subgraphs from the main graph using a breadth-first search (BFS) sequence. A dataset of \(N= 500\) connected subraphs is built by repeating \(N\) times: 1) a number of nodes is sampled from a normal distribution of given mean \( \mu \) and variance \(\sigma^2\), \(n \sim \mathcal{N}(\mu, \sigma^2) \); 2) a BFS ordering of the nodes of the main graph is generated as described in Sec.~\ref{subsubsec.AdjEncode} (randomizing first the nodes numbers (labels)); and 3) the subgraph corresponding to the first \(n\) nodes is extracted, and the node coordinates are shifted such that the mean position is the origin. Note that BFS ensures the selection of the nearest neighbors of the starting node. It starts at the tree root and explores all nodes at the current depth before moving on to the nodes at the next depth level. Therefore, the extracted subgraphs best preserve the local topology. Moreover, centralizing each graph on the origin is not a limitation, since a translation does not affect the geometry of the network. Using the same procedure, validation or testing sets can also be generated.

Then, given the dataset, the two deep learning networks, GraphRNN and G-DDPM that learn respectively the topology and graph node features, are trained independently. Note that to feed the GraphRNN during the training phase, a BFS ordering strategy is applied to the input graphs, then each graph of the dataset is encoded with different adjacency matrices (with a fixed number \(m\) of columns), which somehow increases the size of the dataset (see Sec.~\ref{subsubsec.AdjEncode}). To train the GraphRNN, the number of epochs is set to \( 5 \cdot (\mu + 3\sigma) \) reflecting the complexity of the topology. For training the G-DDPM, each feature \(s\) is first rescaled by applying \(s\mapsto (s-\mu_s)/\sigma_s\), where \(\mu_s\) and \(\sigma_s\) are respectively the mean and the standard deviation of the considered features over the whole dataset. This rescaling helps the diffusion process to reach a Gaussian noise for each feature. The number of timesteps is set to \num{2400} and the number of epochs is set to \num{1201} for training the G-DDPM.

Once trained, these two networks are used to generate a set of \(N=500\) graphs: first, the GraphRNN is employed to generate the topology of the graphs, and nodes features are then generated on these graphs, starting with a Gaussian noise for each feature and applying the G-DDPM, followed by the inverse mapping \(s\mapsto \mu_s + \sigma_s\cdot s\) to each feature \(s\), to get the final features. For the first case study (Ox Bel Ha), only two features are considered, the \(x\) and \(y\) coordinates of the nodes, whereas for the second case study (Sakany), five node features are considered: the three spatial coordinates plus the cross-section width and height (measures provided in the main graph).

\subsection{Coastal–Karst : Ox Bel Ha}
\label{SubSec.OxBelHa}

\begin{figure}[!ht]
    \centering
    \includegraphics[width=0.75\linewidth]{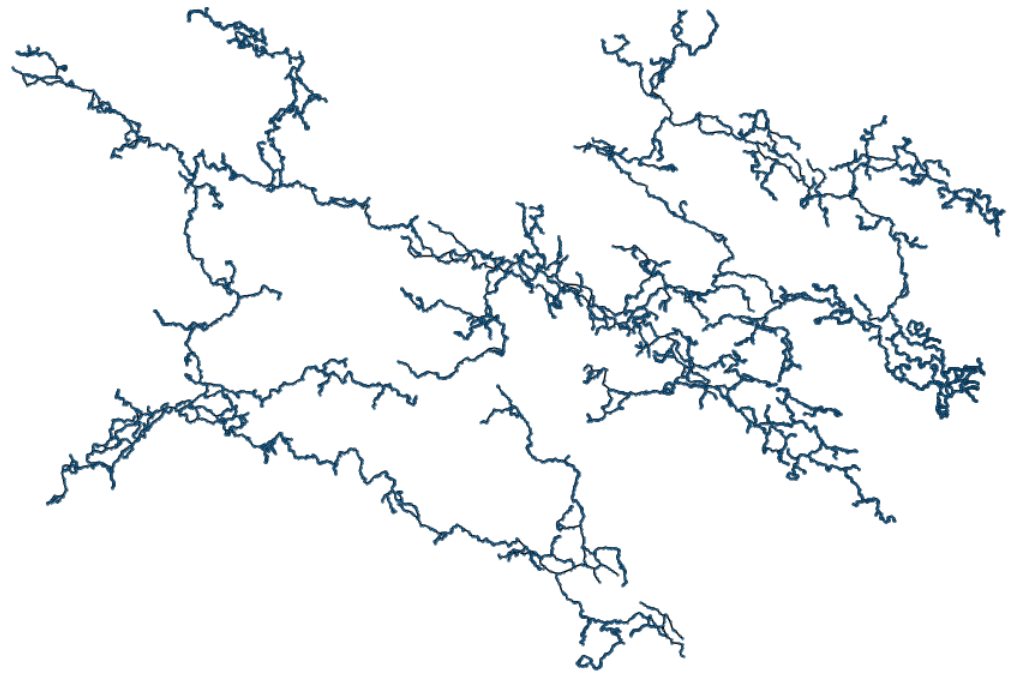}
    \caption{Main graph of the Ox Bel Ha karst system, in two dimension (the vertical dimension is omitted, because only very low variations are observed), \num{10098} nodes, \num{10098} edges.}
    \label{fig.OxBelHa_main_graph}
\end{figure}

Ox Bel Ha, one of the longest cave systems in Mexico, is also among the largest underwater caves in the world. With over 270 kilometers of surveyed submerged passages, the system includes more than 143 cenote entrances and connects to five known ocean discharge points. Spanning an area of over 52 square kilometers, it represents a vast inland hydrological network. The densest cave development is concentrated in a central maze of interconnected drainage conduits, spanning approximately 38 square kilometers. For a comprehensive geological review of the underwater caves of the Yucatán Peninsula, including Ox Bel Ha, refer to \citeA{Coke2019}. Figure~\ref{fig.OxBelHa_main_graph} illustrates the part of the Ox Bel Ha network \cite{qrss_quintana_roo_speleological_survey_ox_2024} used in our tests, it comprises \num{10098} nodes interconnected with \num{10098} edges.

\begin{figure}[!ht]
    \centering
    \includegraphics[width=0.95\linewidth]{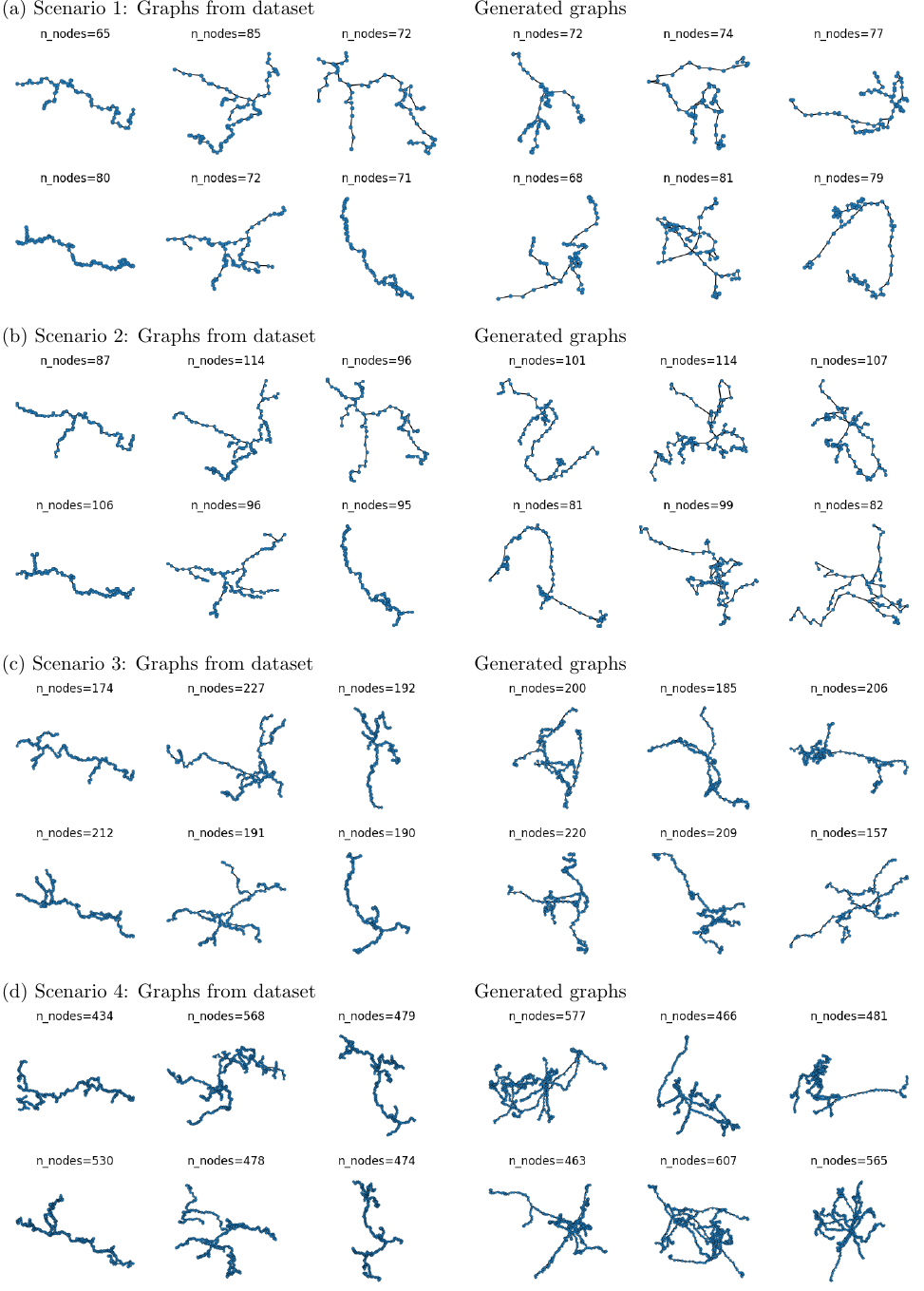}
    \caption{First graphs of datasets (left column) and first generated graphs (right column), where the dataset is built with a number of nodes following the distribution (a) \( \mathcal{N}(75, 7.5^2) \), (b) \( \mathcal{N}(100, 10^2) \), (c) \( \mathcal{N}(200, 20^2) \), and (d) \( \mathcal{N}(500, 50^2) \).}
    \label{fig.OxBelHa_first_graphs_dataset_and_genset}
\end{figure}

To evaluate the flexibility of the proposed approach, four different datasets are used, where the number of nodes in the extracted subgraphs follows respectively the Gaussian distributions \( \mathcal{N}(75, 7.5^2) \), \( \mathcal{N}(100, 10^2) \), \( \mathcal{N}(200, 20^2) \), and \( \mathcal{N}(500, 50^2) \). Figure~\ref{fig.OxBelHa_first_graphs_dataset_and_genset} shows 6 subgraphs from the four datasets in the left column and 6 generated graphs for the four scenarios in the right column.

In the left column of the Fig.~\ref{fig.OxBelHa_first_graphs_dataset_and_genset}, one can observe that the local topology may differ significantly with fewer nodes, and as the number of nodes increases, the network more closely resembles the network maze of the main graph in Fig.~\ref{fig.OxBelHa_main_graph}. One can visually compare the similarities and differences between the training sets (left column) and the generated graphs (right column). First, similar structures can be observed such as labyrinth-like patterns, characterized by abrupt 90-degree turns. However, some generated graphs still exhibit anomalies, such as noise or poorly reproduced configurations, especially in the last scenario. These aberrations reflect the difficulty of GraphRNN combined with G-DDPM in capturing the full complexity of the training graphs.

\begin{figure}[!ht]
    \centering
    \includegraphics[width=0.95\linewidth]{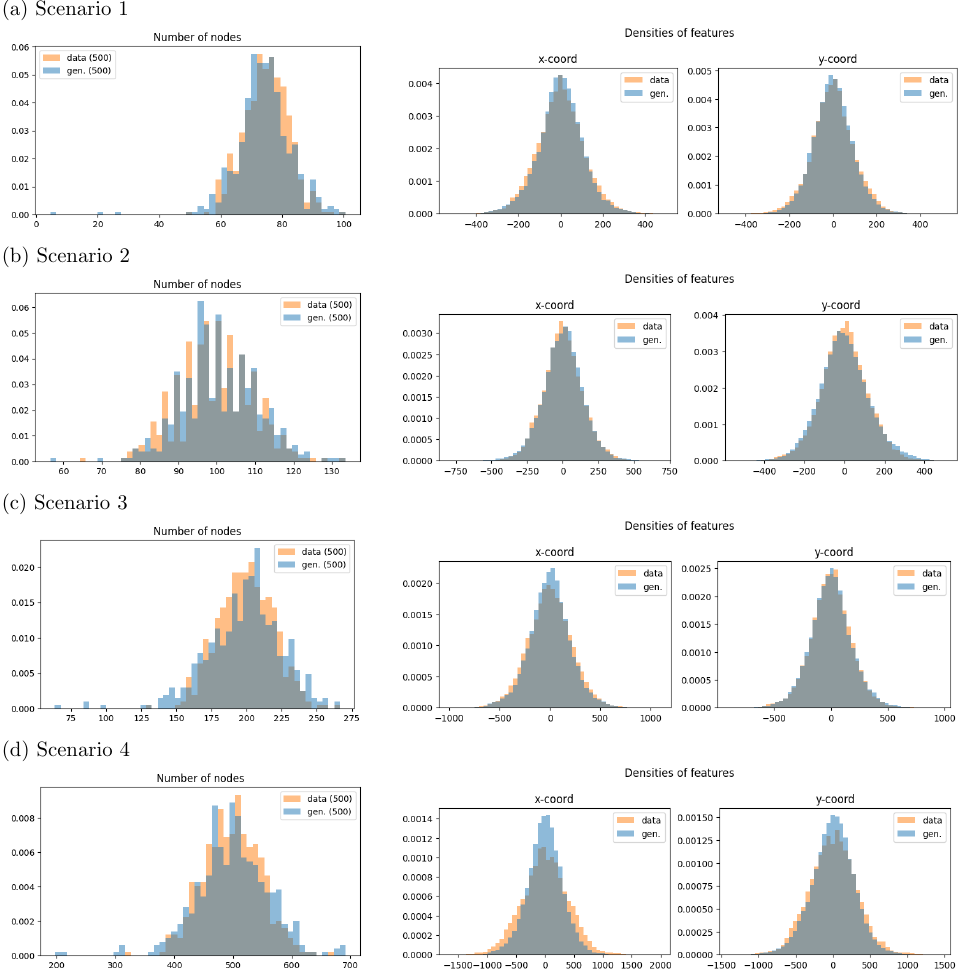}
    \caption{Histogram of number of nodes (left), and \(x\) coordinate (middle column) and \(y\) coordinate (right column) of nodes for all graphs in the dataset (orange) and all generated graphs (blue) for the four scenarios (each row).}
    \label{fig.OxBelHa_stats_marginal}
\end{figure}

In the left columns of Fig.~\ref{fig.OxBelHa_stats_marginal}, the histograms of the number of nodes in the \num{500} graphs of the dataset (orange) and the \num{500} generated graphs (blue) shows that the GraphRNN learns well the number of nodes in the networks, for the four scenarios. In the two other columns, the marginal histograms of the node features (\(x\) and \(y\) coordinates), computed over all graphs in the data sets (orange) and all generated graphs (blue) are displayed. The distributions from the dataset and from the generated graphs are quite similar, although the dispersion is smaller for the generated graphs in the last scenario, involving a larger number of nodes and making the task of the G-DDPM harder.

\begin{figure}[!ht]
    \centering
    \includegraphics[width=0.95\linewidth]{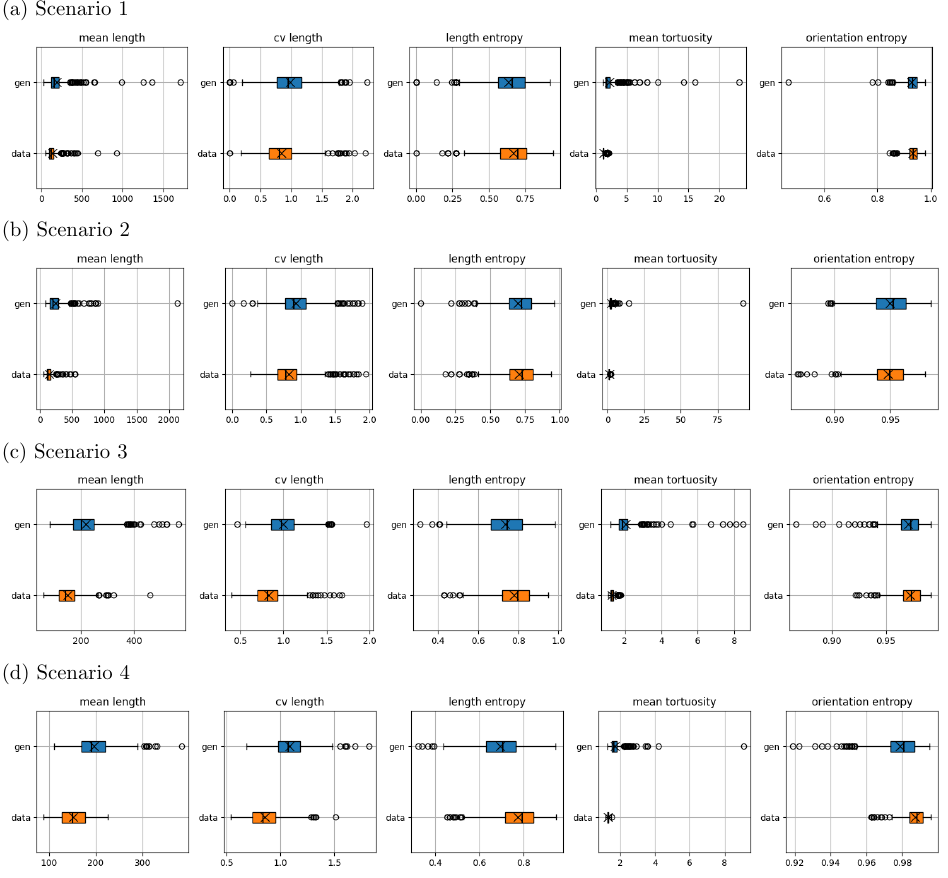}
    \caption{Geometry metrics (from left to right: `mean length, `cv length', `length entropy', `mean tortuosity', `orientation entropy', see Table~\ref{Tab.KarstNetStats}) for graphs in the dataset (orange) and generated graphs (blue), for the four scenarios.}
    \label{fig.OxBelHa_stats_geometry}
\end{figure}

\begin{figure}[!ht]
    \centering
    \includegraphics[width=0.95\linewidth]{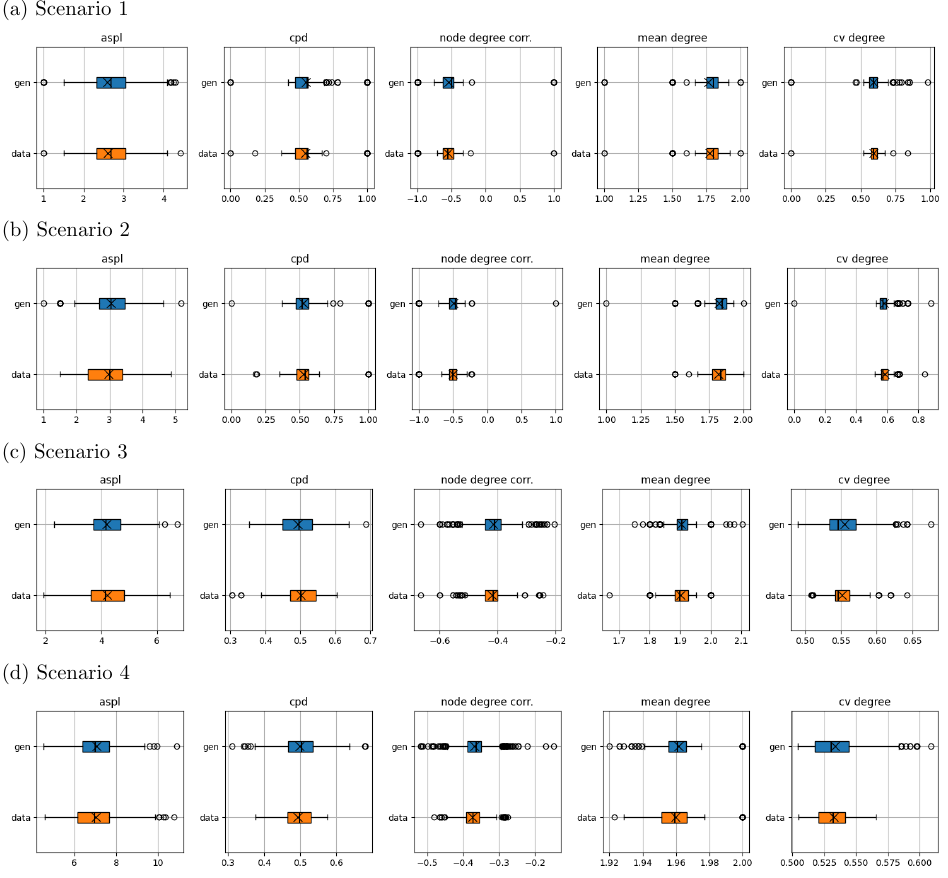}
    \caption{Topology metrics (from left to right: `aspl', `cpd', `node degree corr.', `mean degree', `cv degree', see Table~\ref{Tab.KarstNetStats}) for graphs in the dataset (orange) and generated graphs (blue), for the four scenarios.}
    \label{fig.OxBelHa_stats_topology}
\end{figure}

The ten statistical metrics presented in Sec.~\ref{Sec.Metrics} are computed for every individual graph of the dataset and every individual generated graph, for the four scenarios. The resulting boxplots for the five geometry metrics are presented in Fig.~\ref{fig.OxBelHa_stats_geometry}, and for the five topology metrics in Fig.~\ref{fig.OxBelHa_stats_topology}.

Features associated with the geometries (Fig.~\ref{fig.OxBelHa_stats_geometry}) are generally representative of the subgraph dataset, except for the tortuosity metric. Specifically, the tortuosity is consistently higher for the generated subgraphs compared to the training set. A high tortuosity means that a conduit made of several nodes with degree 2 is not linear, but rather convoluted and winding, with numerous deviations from a straight path. This results in a more complex structure, where the path between nodes is longer than the straight-line distance between them. In our case, this may be caused by noisy coordinate data generated by G-DDPM, or a difficulty in generating subgraphs by G-DDPM using GraphRNN's output. Note that the topology predicted by GraphRNN is unseen during G-DDPM training. However, the extremes in the tortuosity metric are reduced as the number of nodes in the distribution increases, but more outliers are still observed for the generated graphs. For the topology metrics (those marked with a * in Table~\ref{Tab.KarstNetStats}) (Fig.~\ref{fig.OxBelHa_stats_topology}), the boxplots show very little difference. Although some extreme values can be observed, the mean and median statistics are very similar. In other words, our generative graph neural network is capable of capturing the system's topology quite well, but it struggles with precision in the geometry, specifically in the spatial representation.

\clearpage
\subsection{3D–Karst : Sakany}
\label{SubSec.Sakany}

Sakany is characterized as a complex and multi-phase phreatic network located in the Pyrenees (Qui\'e, Ari\`ege, France), with approximately 5 kilometers of developing cavities. Currently, speleologists remain perplexed by the distinct morphologies of the conduits—featuring winding tubes and sub-vertical passages—that are unique, as well as the absence of vadose forms, which complicates its classification. The Sakany network appears as a labyrinthine tangle of conduits, with most entrances located on the southern side. The channels are predominantly sub-horizontal or vertical, forming a complex 3D network. A speleogenetic review of this system can be found by referring to \citeA{cassou2007}.

\begin{figure}[!ht]
    \centering
    \includegraphics[width=0.95\linewidth]{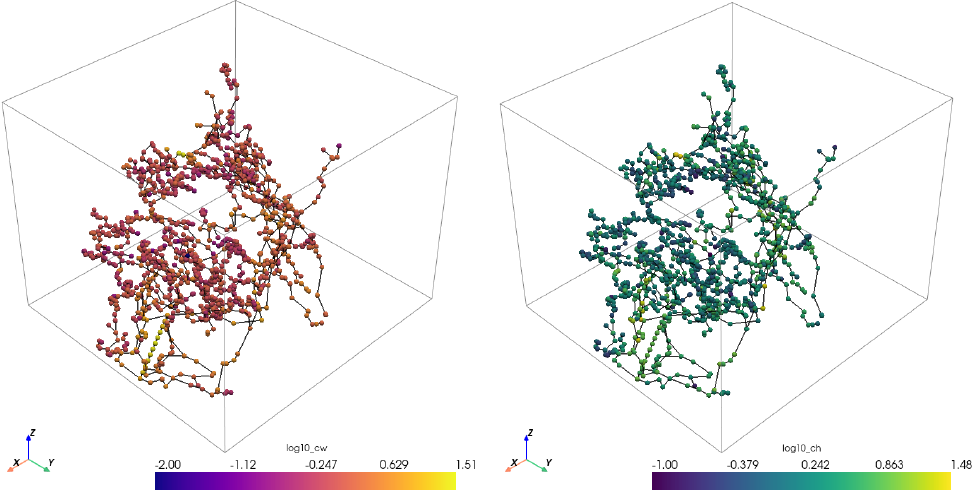}
    \caption{Main graph of the Sakany karst system, \num{1424} nodes, \num{1463} edges; node features are represented: logarithm in base \(10\) of cross-section width (`log10\_cw', left) and height (`log10\_ch', right).}
    \label{fig.Sakany_main_graph}
\end{figure}

The largest connected part of this karst system \cite{groupe_de_recherches_et_dactivites_speleologiques_de_lourdes_base_2024} is considered, with cross-section width and height at each node, measures of the distances between ``opposite'' walls of the conduit. Note that in the original network, these measures were unknown at a very few nodes, or equal to zeros for entries of the cave; in such situations, the value is set to the mean over the neighboring nodes. Then the features named `log10\_cw' and `log10\_ch' are retrieved by taking the logarithm in base 10 of the cross-section width and height respectively. Using the logarithm implies that negative feature values are allowed. The main system, composed of \num{1424} nodes and \num{1463} edges, is shown with these two features in Fig.~\ref{fig.Sakany_main_graph}.

\begin{figure}[!ht]
    \centering
    \includegraphics[width=0.95\linewidth]{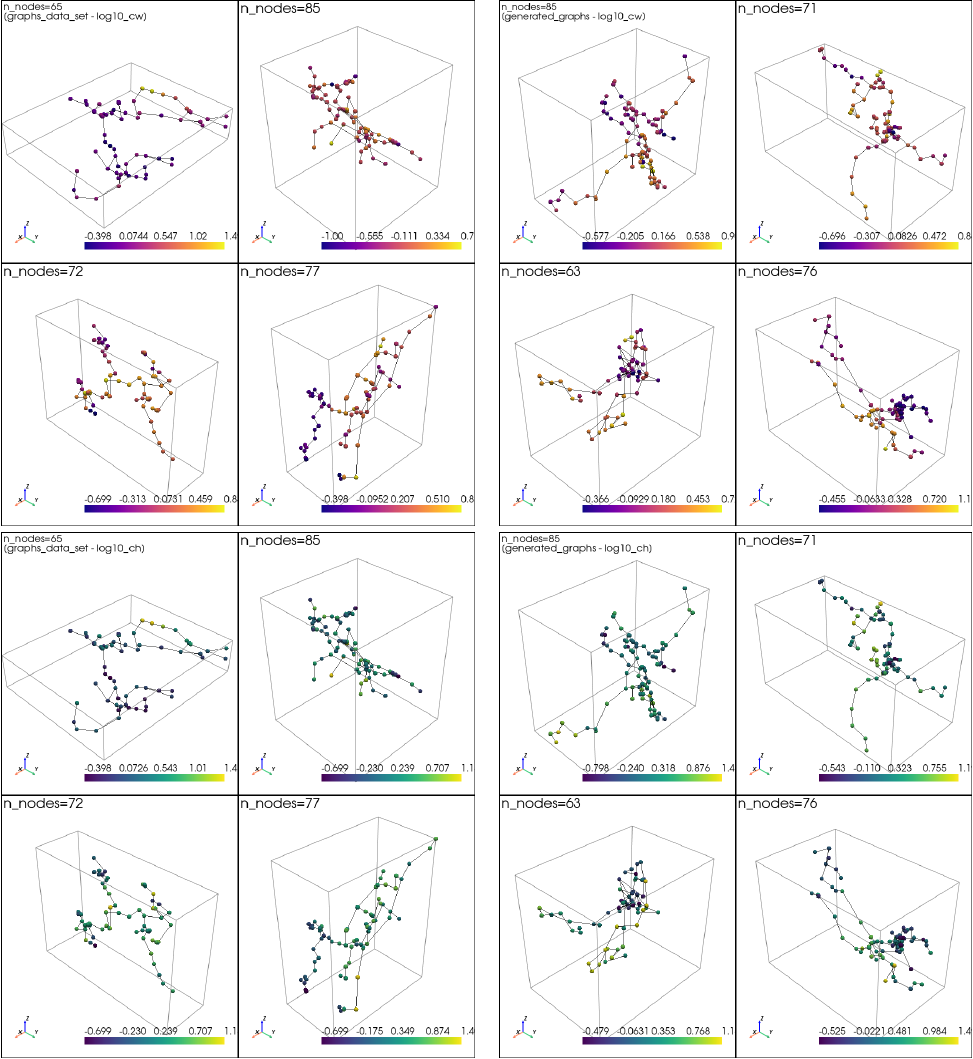}
    \caption{First graphs of the dataset (left column) and first generated graphs (right column), node feature `log10\_cw' (top row), and `log10\_ch' (bottom row).}
    \label{fig.Sakany_first_graphs_dataset_and_genset}
\end{figure}

The dataset used to train the deep neural networks is built by randomly extracting \num{500} connected subgraphs with a number of nodes drawn in the Gaussian distribution \( \mathcal{N}(75, 7.5^2) \). First graphs of the dataset and first generated graphs are displayed in Fig.~\ref{fig.Sakany_first_graphs_dataset_and_genset}. Based on a visual inspection, the results seem to reasonably capture the structures from the dataset.

\begin{figure}[!ht]
    \centering
    \includegraphics[width=0.95\linewidth]{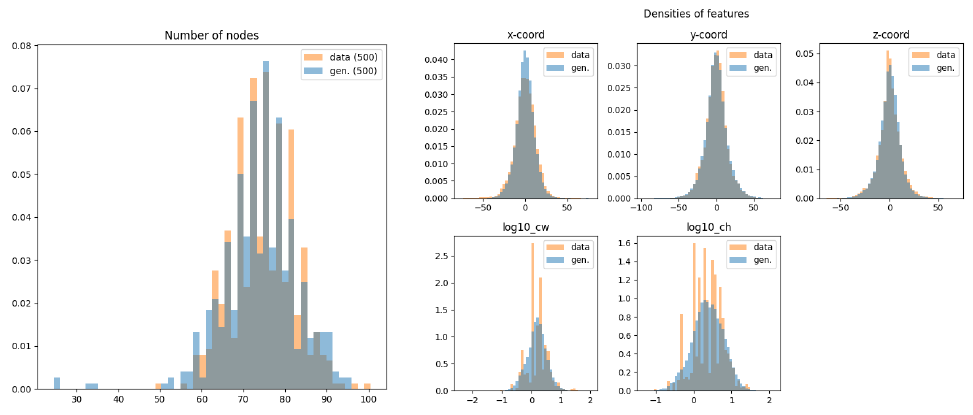}
    \caption{Histogram of number of nodes (left), and each node features (\(x, y, z\) coordinates and `log10\_cw', `log10\_ch', right) for all graphs in the dataset (orange) and all generated graphs (blue) for the four scenarios (each row).}
    \label{fig.Sakany_stats_marginal}
\end{figure}

Statistics are computed on the dataset (\num{500} graphs) and on the \num{500} generated graphs. The histograms of the number of nodes and the marginal densities of the five nodes features (three spatial coordinates, and two cross-section measures) are presented in Fig.~\ref{fig.Sakany_stats_marginal}. One can observe that the number of nodes is well-learned by the GraphRNN. The G-DDPM well reproduces the distributions of the spatial coordinates of the nodes, but those of the two extra node features (`log10\_cw' and `log10\_ch') are more different, the distribution of the generated extra features struggle to deviate from a Gaussian. Note that correlation betweeen the features and spatial statistics are not considered in this work. It is important to note that for this case study the main graph from which the dataset is built contains ``only'' \(1424\) nodes. As a consequence, the dataset may not be rich enough to allow an efficient learning.

\begin{figure}[!ht]
    \centering
    \includegraphics[width=0.95\linewidth]{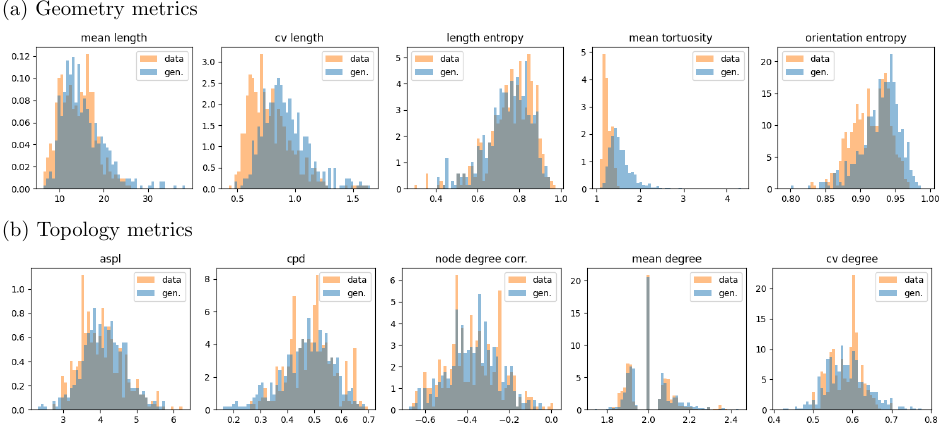}
    \caption{Geometry (a) and topology (b) metrics (see Table~\ref{Tab.KarstNetStats}) for graphs in the dataset (orange) and generated graphs (blue).}
    \label{fig.Sakany_stats_geometry_topology}
\end{figure}

The ten statistical metrics presented in Sec.~\ref{Sec.Metrics} are displayed in Fig.~\ref{fig.Sakany_stats_geometry_topology}. As for the first case study, the geometry metrics are well reproduced, excepted the mean tortuosity and the coefficient of variation (`cv length') for which generated graphs have higher values.

\clearpage
\section{Discussion and Conclusions}
\label{Sec.DisConc}

We explore a novel approach to model karstic systems using deep neural networks. The method follows a two-step process: first, it learns the system's topology using GraphRNN \cite{You2018}, and second, it simulates the spatial structure by generating spatial coordinates and other properties using G-DDPM conditioned on the learned topology by GraphRNN. To increase the diversity of training data, we apply data augmentation, transforming a single main graph into multiple subgraphs, constituting the dataset used for training. This approach is motivated by the limited access to several karst network systems, and allows to provide enough data to feed the neural networks. For a given karstic system under study, represented as a graph, a set of connected subgraphs is extracted randomly with a varying number of nodes, sampled from a Gaussian distribution. This enables the proposed model to generate graphs of varying sizes and capture the natural variability of real-world systems. GraphRNN captures the topology (i.e., it generates a topology with a given number of nodes) and G-DDPM spatially position the generated graph along side other properties.

The proposed approach demonstrates flexibility in graph generation, effectively adapting to karst systems across a range of scales and complexities. Its versatility is evident in the successful modeling of both a 2D underwater cave system and a 3D alpine karst network. Furthermore, the experiments conducted on the two karst conduit network datasets and depicting very different features yielded consistent results, further reinforcing the strengths of the method. As the method is based on statistical learning, the specific karstic formation processes and mechanisms don't matter, and it can be applied to distinct karstic environments.

The statistical metrics used to evaluate an ensemble of generated karstic networks are based on geometric and topological characteristics. Geometry metrics are computed on the complete graph, whereas topology metrics are computed on a simplified graph where essentially nodes of degree 2 are removed. Out of the 10 metrics (5 for geometry and 5 for topology), only the tortuosity, a geometry metric, proves complex to reproduce. This indicates that the proposed methodology can generate an accurate skeleton of the karst but struggles to correctly reproduce the tortuosity of the branches, or the intermediate connections in the skeleton, which are primarily linked to nodes of degree 2. It is important to note that the number of internal nodes with degree 2 in a topographical survey is largely influenced by field conditions and the speleologist's sampling preferences, which can vary significantly across different parts of a cave system \cite{Collon2017}. This variability may complicate the learning process. Furthermore, the tortuosity itself is influenced by the number of points and the way to measure them along the karst conduits. Thus, to make this metrics more meaningful, it could be useful to pre-process the initial mapped network, for instance by smoothing the branches until the angles between two successive edges fall between a maximal predefined value (to avoid to sharp angles).

An intriguing future application of our deep neural network is the prediction of maps for unexplored or inaccessible areas. By generating variable-sized graphs, our network can effectively capture the uncertainty associated with such regions. For instance, in the Ox Bel Ha karst network, certain sections remain unmapped due to human or equipment access limitations. Our method has the potential to propose plausible structures to fill these gaps. However, further research and validation are required, particularly to enable conditioning on existing paths and ensuring realistic extrapolations. This could potentially be addressed by using an equivariant 3D-conditional diffusion model, as proposed by \citeA{Igashov2024} for molecular linker design: given a set of disconnected fragments, their model places missing nodes in between and designs a network incorporating all the initial fragments. However, the transfer of this approach to karst system generation remains uncertain, and further studies are needed.

Another perspective consists in accounting for the surrounding geology in karst system. This require to explore the possibility of incorporating categorical features into G-DDPM, which could reflect geological settings. These categorical features would modify the complexity of the loss function by combining continuous and categorical features simultaneously. Additionally, categorical data could accommodate successive phases of speleogenesis under changing geological and hydrogeological boundary conditions. Therefore, we recall that our approach aims to explore the use of deep neural networks for modeling complex geological mechanisms, such as karstic networks, within a framework where data is limited. We conclude that this research area is entirely new and opens up a wide range of potential research, development, and applications that can enhance our understanding of these geological systems.

\section*{Declaration of Competing Interest}
The authors declare that they have no conflicts of interest.

\section*{Open Research Section}
The code to apply the proposed approach and the data for the Sakany cave study are available in the repository at the following url: https://doi.org/10.5281/zenodo.15090730. Note that Ox Bel Ha data are not freely available.

\acknowledgments

The authors acknowledge funding by the European Union (ERC, KARST, 101071836). Views and opinions expressed are however those of the authors only and do not necessarily reflect those of the European Union or the European Research Council Executive Agency. Neither the European Union nor the granting authority can be held responsible for them.

The authors thank James G. Coke for spending the time preparing and sharing with us the survey data of the Ox Bel Ha karst network, the GEO (Grupo de Exploración Ox Bel Ha), the MCEP (Mexico Cave Exploration Project), the QRSS (Quintana Roo Speleological Survey), the CINDAQ (El Centro Investigador del Sistema Acuífero de Quintana Roo), and all the cave surveyor who explored and mapped the cave over the years. The authors also thank the Sakany cave explorers and all those that contributed to the related dataset and that made it freely available.

The authors would also like to thank Celia Trunz for processing the raw data used in this work.

Dany Lauzon is also grateful to Polytechnique Montréal for funding his postdoctoral research position at the University of Neuchâtel.

\bibliography{biblio}

\end{document}